\documentclass[10pt,twocolumn,letterpaper]{article}
\usepackage{cvpr}
\definecolor{cvprblue}{rgb}{0.21,0.49,0.74}
\usepackage[pagebackref,breaklinks,colorlinks,allcolors=cvprblue]{hyperref}
\usepackage{tabularx}
\usepackage{array}
\newcolumntype{Y}{>{\centering\arraybackslash}X}
\newcolumntype{Z}{>{\raggedleft\arraybackslash}X}
\usepackage{arydshln}
\usepackage{colortbl}
\usepackage{algorithm,algorithmic}

\title{DeRA: Decoupled Representation Alignment for Video Tokenization}

\author{
Pengbo Guo$^{1,2}$ \quad Junke Wang$^{3}$ \quad Zhen Xing$^{3}$ \quad Chengxu Liu$^{1}$ \\  Daoguo Dong$^{3}$ \quad Xueming Qian$^{1}$ \quad Zuxuan Wu$^{2,3}$\\ 
$^1$Xi'an Jiaotong University \\
$^2$Shanghai Innovation Institute \\
$^3$Institute of Trustworthy Embodied AI, Fudan University \\
}

\begin{document}
\maketitle

\begin{abstract}
This paper presents DeRA, a novel 1D video tokenizer that decouples the spatial-temporal representation learning in video tokenization to achieve better training efficiency and performance. Specifically, DeRA maintains a compact 1D latent space while factorizing video encoding into appearance and motion streams, which are aligned with pretrained vision foundation models to capture the spatial semantics and temporal dynamics in videos separately. To address the gradient conflicts introduced by the heterogeneous supervision, we further propose the Symmetric Alignment-Conflict Projection (SACP) module that proactively reformulates gradients by suppressing the components along conflicting directions. Extensive experiments demonstrate that DeRA outperforms LARP, the previous state-of-the-art video tokenizer by 25\% on UCF-101 in terms of rFVD. Moreover, using DeRA for autoregressive video generation, we also achieve new state-of-the-art results on both UCF-101 class-conditional generation and K600 frame prediction.
\end{abstract}

\section{Introduction}
\label{sec:intro}
Video tokenizers~\cite{yu2023magvit,wang2024omnitokenizer,tian2024reducio,wang2024larp} transform raw video inputs into compact latent representations, serving as a crucial interface for generative models~\cite{jia2025principles,yao2025reconstruction,tian2025unigen,mahapatra2025progressive}. Existing designs generally fall into two categories: continuous tokenizers~\cite{kingma2013auto,yao2025reconstruction,wang2025vidtwin} which map inputs to probability distributions, and discrete tokenizers~\cite{van2017neural,esser2021taming,yu2021vector} that quantize visual signals into sequences of latent tokens. Analogous to the tokenization process in language models, discrete video tokenizers project high-dimensional video inputs into a finite vocabulary of token indices. This formulation enables modern language models to perform visual generation as a next-token prediction process~\cite{sun2024autoregressive,wang2024emu3,wang2025simplear}, effectively leveraging their strong sequence modeling capabilities and favorable scaling behavior for high-fidelity visual synthesis.

Typically, existing discrete video tokenizers first patchify an input video into spatial-temporal tubelets and then encode them using convolutional layers~\cite{ge2022long,yu2023language} or transformer blocks~\cite{villegas2022phenaki,wang2024omnitokenizer,tan2025sweettok} to obtain compact latent representations. These approaches excel at preserving local visual details and achieve strong reconstruction performance on standard benchmarks~\cite{soomro2012dataset,carreira2018short}. However, this fixed grid design leads to a quadratic growth of token count as spatial resolution and temporal length increase~\cite{yu2024image,bachmann2025flextok}. To overcome this inefficiency, recent works have explored one-dimensional (1D) image tokenization~\cite{yu2024image,bachmann2025flextok}, which compresses an image into a sequence of latent queries, dramatically shortening the sequence length~\cite{li2025learning,xiong2025gigatok} with minimal sacrifice in reconstruction quality. Extending this paradigm to video, several works, exemplified by LARP~\cite{wang2024larp}, learn a set of 1D queries to compress video clips.

\begin{figure}[!tb]
\centering
\includegraphics[width=0.48\textwidth]{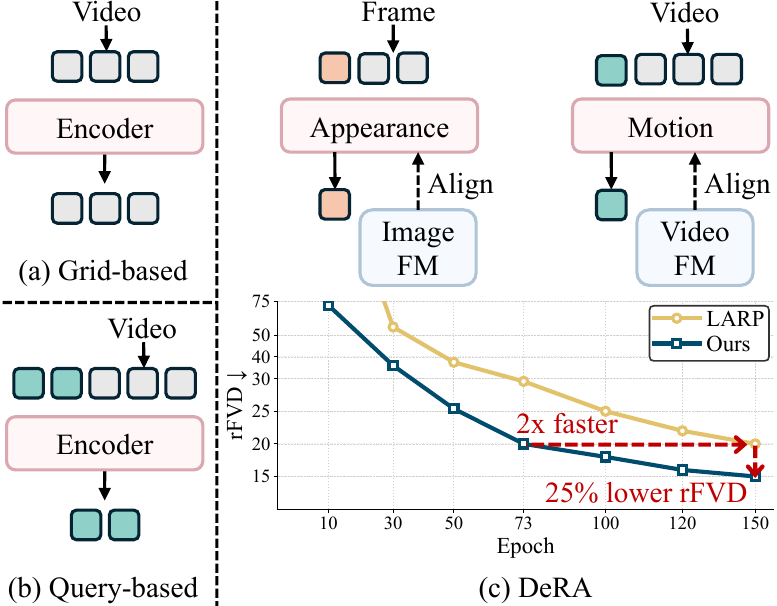}
\vspace{-0.2in}
\caption{Comparison of video tokenization approaches. (a) Grid-based tokenizer; (b) Query-based 1D tokenizer; (c) DeRA aligns decoupled representations to the corresponding foundation model (FM), yielding 2$\times$ faster convergence and better performance.}
\label{fig:intro}
\vspace{-0.1in}
\end{figure}

While 1D video tokenizers can achieve comparable reconstruction performance with far fewer tokens by capturing global semantics better~\cite{wang2024larp}, existing methods often suffer from low training efficiency. For instance, LARP~\cite{wang2024larp} takes more than twice as many training steps as MAGVIT-v2 to converge~\cite{luo2024open}. In this paper, we demonstrate that the inefficiency largely stems from their encoding mechanism, \ie, these models entangle the spatial and temporal modeling of video inputs within a single transformer backbone, which complicates optimization and limits the model to effectively capturing fine-grained motion dynamics and spatial details. We argue that a more effective paradigm is to explicitly decouple the spatial-temporal learning of video tokenization with dedicated pathways, as shown in Fig.~\ref{fig:intro}.

With this in mind, we present DeRA, a novel 1D video tokenizer that introduces explicit supervision for spatial-temporal decoupling in video tokenization. We factorize the video encoding process into two streams: an appearance stream that focuses on static visual details from the first frame, and a motion stream that processes the entire video clip to capture dynamic variations. We employ independent query sets for each stream to extract complementary spatial and temporal representations. To enforce representation disentanglement between them, DeRA aligns the appearance queries with representations from a powerful image foundation model and the motion queries with those from a motion-centric video foundation model. Furthermore, to mitigate the gradient conflicts arising from these heterogeneous supervisory signals, we propose the Symmetric Alignment-Conflict Projection (SACP) module, which reformulates competing gradients into a consensus direction that benefits both alignment objectives. Experiments on UCF-101~\cite{soomro2012dataset} and Kinetics-600~\cite{carreira2018short} demonstrate that our DeRA achieves a superior reconstruction rFVD of 15 and sets new state-of-the-art results on class-conditional generation and frame prediction benchmarks.

In summary, our work makes the following key contributions:
\begin{itemize}
\item We propose DeRA, a novel 1D video tokenizer that achieves explicit spatial-temporal decoupling by aligning representations with multiple vision foundation models.
\item We introduce SACP, a gradient reformulation module that mitigates multi-objective conflicts during decoupled alignment, stabilizing optimization and improving representation consistency.
\item Extensive experiments across multiple benchmarks demonstrate that DeRA consistently outperforms existing video tokenizers by clear margins. When integrated into AR generation frameworks, it further achieves state-of-the-art performance on UCF-101 class-conditional generation and Kinetics-600 frame prediction tasks.
\end{itemize}

\section{Related Work}
\label{sec:relatedwork}

\subsection{Video Tokenizers}
Building upon the success of image tokenizers, existing video tokenization approaches adapt similar architectures to process 3D spatial-temporal data. Notable models such as TATS~\cite{ge2022long}, CViViT~\cite{villegas2022phenaki}, and MAGVIT-v2~\cite{yu2023language} employ grid-based methods, treating video as a volume of patch embeddings. OmniTokenizer~\cite{wang2024omnitokenizer} further extends this approach with a unified model capable of tokenizing both images and videos. However, these grid-based methods often result in excessively long and redundant token sequences and overlook the high-level semantics, which impose significant challenges for subsequent AR modeling. Recently, a paradigm shift has occurred towards enhancing the efficiency and semantic richness of tokenizers to benefit downstream generation. Pioneered in the image domain by TiTok~\cite{yu2024image}, which abandons the grid-based approach by directly encoding an entire image into a compact, 1D sequence of tokens. LARP~\cite{wang2024larp} extends this paradigm to the video domain, introducing a prior model to align the token order during training with the sequential nature of AR generation. Despite these advances, existing video tokenizers still suffer from time-consuming training, and struggle to simultaneously preserve static details and dynamic motion.

\begin{figure*}[!tb]
    \centering
    \includegraphics[width=\textwidth]{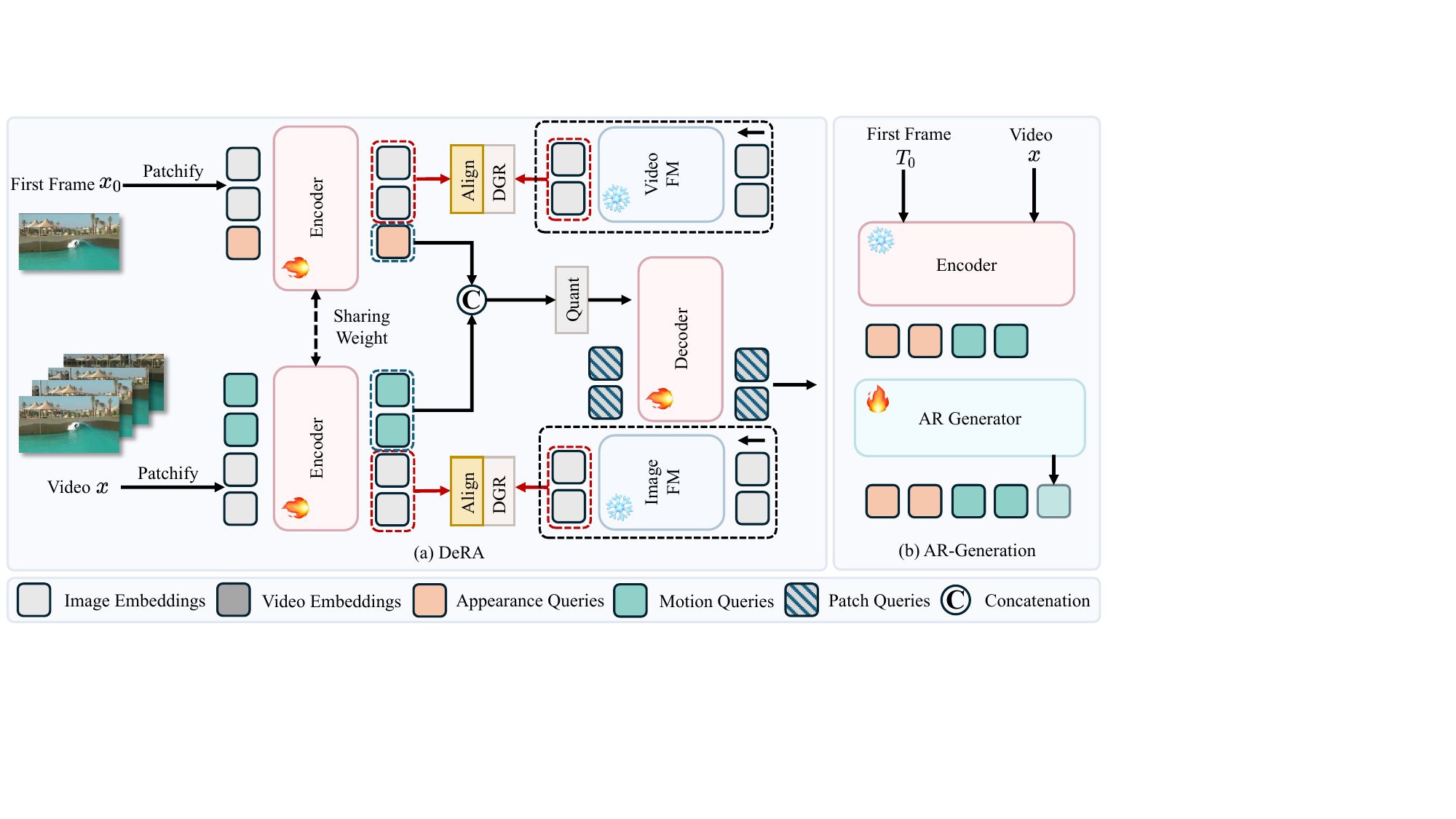}
    \vspace{-0.2in}
    \caption{Overview of our method. DeRA decouples video into the appearance stream and motion stream via a shared encoder. During training, we align appearance and motion latents to the frozen image and video foundation models, respectively. The Symmetric Alignment-Conflict Projection (SACP) module reformulates the conflicting gradients before updating. The two latents are concatenated and quantized to produce discrete tokens that drive the decoder for reconstruction. During inference, external models and regularizations are removed.}
    \label{fig:overview}
\end{figure*}

\subsection{Visual Generation Modeling with Representation Alignment}
Distilling knowledge from large-scale foundation models~\cite{oquab2023dinov2,simeoni2025dinov3,wang2024internvideo2} has emerged as a powerful approach to circumvent data scarcity and accelerate the training of generative models. This paradigm was first introduced by REPA~\cite{yu2024representation}, which regularizes the intermediate features of the diffusion model to align with representations from pretrained vision models, accelerating convergence and improving generation quality. REPA-E~\cite{leng2025repa} enables end-to-end fine-tuning by demonstrating the backpropagation of the REPA loss to jointly optimize both the VAE tokenizer~\cite{kingma2013auto} and the diffusion model~\cite{ho2020denoising,nichol2021improved}. REG~\cite{wu2025representation} further refines this process by entangling a noised class token with image latents, ensuring persistent guidance throughout the denoising process. VideoREPA~\cite{zhang2025videorepa} introduces a soft alignment loss designed specifically for fine-tuning pretrained video diffusion models, distilling implicit understanding of temporal dynamics and physics. Building on these insights, we posit that a more effective strategy is to harness the high-quality features from multiple foundation models to guide explicitly decoupled video tokenization.

\subsection{Autoregressive Video Generation}
Inspired by the success of large language models, AR-based models have emerged as a popular paradigm for visual synthesis, offering a promising direction towards unified multimodal modeling and robust generalization over long temporal horizons~\cite{jia2025principles,wangomnigen}. Early attempts in AR-based generation model images as sequences of pixels~\cite{van2016pixel}, achieving satisfactory results but suffering from prohibitive sequence lengths that render long-duration video generation impractical. The advent of discrete visual tokenization (e.g., VQ-VAE~\cite{van2017neural}) reframes visual generation as a next-token prediction paradigm in a compact latent space. Building upon this, a surge of video AR models demonstrate increasingly impressive performance~\cite{yan2021videogpt,ge2022long,wang2024emu3}. These works further refine the AR transformer to model complex temporal dynamics and scale to longer sequences. With the advancement of generative models, the demand for an efficient, high-fidelity video tokenizer has become a critical frontier, as it directly determines the sampling efficiency and fidelity~\cite{yu2023language,wang2024omnitokenizer}.

\section{Methodology}
\label{sec:method}
We aim to decouple spatial-temporal representation learning in video tokenization to boost training efficiency and performance. To this end, we design a dual-stream architecture that separately captures appearance and motion cues in Sec.~\ref{sec:archi}. To explicitly supervise the decoupled learning, we propose the Decoupled Representation Alignment framework in Sec.~\ref{sec:dra}. We further introduce the Symmetric Alignment-Conflict Projection module to resolve the gradient conflicts arising from simultaneously aligning with heterogeneous objectives in Sec.~\ref{sec:sacp}. The overall framework of DeRA is shown in Fig.~\ref{fig:overview}.

\subsection{Decoupled 1D Video Tokenizer}
\label{sec:archi}
DeRA follows a concise ``encoder-quantizer-decoder'' tokenization architecture. The factorized dual-stream encoder employs distinct appearance and motion latent queries to process the input video and its first-frame patch embeddings, producing compact latent tokens. These tokens are subsequently vector-quantized into discrete representations, which the decoder then combines with its latent queries to reconstruct the original video sequence.

\noindent\textbf{Patchify.} Given an input video $\mathbf{x} \in \mathbb{R}^{T \times H \times W \times 3}$, where $T$ is the number of frames and $H \times W$ denotes the spatial resolution. We use the first frame $\mathbf{x}_0$ as the input to the appearance stream to capture the static content of the video~\cite{wang2024omnitokenizer}. As for the motion stream, we feed the full clip $\mathbf{x}$ to model the temporal dynamics. Specifically, both $\mathbf{x}_0$ and $\mathbf{x}$ are split into non-overlapping patches, with a patch size of $p \times p$ and $t \times p \times p$, respectively. Then we use two linear layers to obtain the corresponding projected patch embeddings $\mathbf{E}_s\in \mathbb{R}^{L_s \times d}$ and $\mathbf{E}_t\in \mathbb{R}^{L_t \times d}$, where $L_{s}=\frac{H}{p} \times \frac{W}{p}$ and $L_{t}=\frac{T}{t} \times \frac{H}{p} \times \frac{W}{p}$, $d$ is the hidden dimension.

\noindent\textbf{Encoder and decoder.} We adopt the dual-stream 1D tokenizer~\cite{yu2024image,wang2024larp} as the base architecture of DeRA, which uses two sets of learnable latent queries to capture the spatial and temporal cues from the frame and video patch embeddings, respectively. For simplicity and scalability, we employ vanilla transformer blocks for both the encoder and decoder. Notably, both streams share the same encoder to ensure computational efficiency. 

To achieve dual-stream tokenization, we introduce two distinct sets of learnable queries: appearance queries $\mathbf{Q}_a \in \mathbb{R}^{L_a \times d}$ and motion queries $\mathbf{Q}_m \in \mathbb{R}^{L_m \times d}$, where $L = L_a + L_m$ is the total number of final discrete tokens. For each stream, we concatenate the queries with corresponding patch embeddings and pass them through the encoder $\mathcal{E}$:
\begin{equation}
\label{equ:enc}
\begin{gathered}
\mathbf{Z}_a\!\parallel\!\mathbf{E}_a=\mathcal{E}(\mathbf{Q}_a \parallel \mathbf{E}_s),\\
\mathbf{Z}_m\!\parallel\!\mathbf{E}_m=\mathcal{E}(\mathbf{Q}_m \parallel \mathbf{E}_t),\\
\end{gathered}
\end{equation}
where $\|$ denotes the concatenation operation, $\mathcal{E}$ is the shared encoder, $\mathbf{Z}_a$ and $\mathbf{Z}_m$ are the output appearance and motion queries, and $\mathbf{E}_a$ and $\mathbf{E}_m$ are the corresponding output embeddings. The queries from both streams are concatenated and quantized using the vector quantizer~\cite{van2017neural} $\text{Quant}(\cdot)$:
\begin{equation}
\label{equ:quant}
\mathbf{y}=\operatorname{Quant}(\mathbf{Z}_a \parallel \mathbf{Z}_m),
\end{equation}
where $\mathbf{y}\in \mathbb{R}^{L \times d_z}$ denotes the quantized discrete tokens.

Concurrently, the decoder $\mathcal{D}$ concatenates the encoded tokens $\mathbf{y}$ with the decoder query embeddings $\mathbf{Q_d}\in \mathbb{R}^{L_d \times d}$, to get the final output $\mathbf{Z}_d$. Then, the first $L_d$ tokens of $\mathbf{Z}_d$ are reshaped to reconstruct the video $\mathbf{x}_{r} \in \mathbb{R}^{T \times H \times W \times 3}$:

\begin{equation}
\begin{split}
    \mathbf{Z}_d=\mathcal{D}({\mathbf{Q}_d} \| \mathbf{y}), \quad \mathbf{x}_{r} = \text{Reshape}(\mathbf{Z}_d[1:L_d]).
\end{split}
\end{equation}

\subsection{Decoupled Representation Alignment}
\label{sec:dra}
Building upon the above architecture, we further introduce explicit supervision to the latent space of encoder $\mathcal{E}$ to effectively decouple the representation learning of DeRA. Rather than aligning to a single compromised foundation model, we separately align the representations of each stream with a specialized expert~\cite{simeoni2025dinov3,wang2024internvideo2}. Specifically, DeRA aligns the latent embeddings $\mathbf{E}_a, \mathbf{E}_m$ of the encoder (in~\cref{equ:enc}) with the output tokens from corresponding frozen image and video foundation models. A lightweight two-layer MLP $h_{\phi}$ projects the latents into the target latent space. We align the projected latents and the target tokens of vision foundation models by minimizing their negative cosine similarity:
\begin{equation}
\begin{split}
    & \mathcal{L}_{\text{align}}^{a}
= -\mathbb{E} [ \frac{1}{L_s} \sum_{n=1}^{L_s} \frac{\mathbf{E}_{ifm}^{n} \cdot h_{\phi}(\mathbf{E}_{a}^{n})}{\|\mathbf{E}_{ifm}^{n}\|_2 \, \|h_{\phi}(\mathbf{E}_{a}^{n})\|_2}],\\
    & \mathcal{L}_{\text{align}}^{m}
= -\mathbb{E} [ \frac{1}{L_t} \sum_{n=1}^{L_t} \frac{\mathbf{E}_{vfm}^{n} \cdot h_{\phi}(\mathbf{E}_{m}^{n})}{\|\mathbf{E}_{vfm}^{n}\|_2 \, \|h_{\phi}(\mathbf{E}_{m}^{n})\|_2}],
\end{split}
\end{equation}
where $\mathbf{E}_{ifm}$ and $\mathbf{E}_{vfm}$ are the target tokens extracted from the image and video foundation models, respectively. Both alignment losses are incorporated in the training loss:
\begin{equation}
\begin{split}
    & \mathcal{L} = \lambda_a\mathcal{L}_{\text{align}}^{a} + \lambda_m\mathcal{L}_{\text{align}}^{m} + \mathcal{L}_{0},
\end{split}
\end{equation}
where $\lambda_a$ and $\lambda_m$ are the weights of alignment losses. $\mathcal{L}_{0}$ is composed of $\ell_1$ reconstruction loss, LPIPS perceptual loss~\cite{zhang2018unreasonable}, GAN loss~\cite{goodfellow2014generative}, and VQ loss~\cite{van2017neural} following~\cite{wang2024larp}. During inference, we only keep the original tokenizer and discard other auxiliary regularization modules. Detailed experiments and discussions can be found in Sec.~\ref{sec:abl}.

\begin{table*}[!tb]
    \centering \setlength{\tabcolsep}{9pt}
    {
        \begin{tabular}{lccccccc}
        \toprule
        \textbf{Method} & \multicolumn{2}{c}{\textbf{Params}} & \textbf{Codebook} & \textbf{Tokens} & \textbf{rFVD}$\downarrow$ & \multicolumn{2}{c}{\textbf{gFVD}$\downarrow$} \\
         \cmidrule{2-3} \cmidrule{7-8}
         & \textbf{Tokenizer}  & \textbf{Generator} &  &  & & K600  & UCF-101  \\
        \midrule
        \multicolumn{7}{l}{\textit{Diffusion-based generative models with continuous video tokenizers}} \\
        \hdashline 
        \addlinespace
        VideoFusion~\cite{luo2023videofusion} & - & 2B & $\infty$ & - & - & - & 173 \\
        HPDM~\cite{skorokhodov2024hierarchical} & - & 725M & $\infty$ & - & - & - & 66 \\
        \midrule
        \multicolumn{7}{l}{\textit{MLM generative models with discrete video tokenizers}} \\
        \hdashline 
        \addlinespace
        MAGVIT-MLM~\cite{yu2023magvit} & 158M & 306M & 1024 & 1024 & 25 & 9.9  &  76 \\
        MAGVIT-v2-MLM~\cite{yu2023language} & - & 307M & 262144 & 1280 & \textbf{8.6} & \underline{4.3} & {58} \\
        \midrule
        \multicolumn{7}{l}{\textit{AR generative models with discrete video tokenizers}} \\
        \hdashline 
        \addlinespace
        CogVideo~\cite{hong2022cogvideo} & - & 9.4B & - & 2065 & - & 109.2 & 626 \\
        TATS~\cite{ge2022long}           & 32M & 321M &  16384 & 1024 & 162 & - & 332 \\
        MAGVIT-AR~\cite{yu2023magvit}    & 158M & 306M & 1024 & 1024 & 25 & - & 265 \\
        MAGVIT-v2-AR~\cite{yu2023language} & - & 840M & 262144 & 1280 & \textbf{8.6} & - & 109 \\
        OmniTokenizer~\cite{wang2024omnitokenizer} & 82.2M & 650M & 8192 & 1280 & 42 & 32.9 & 191 \\
        ElasticTok~\cite{yan2024elastictok} & 200M & - & 64000 & 2048 & 93 & - & - \\
        AdapTok~\cite{li2025learning} & 195M & 633M & 8192 & 2048 & 28 & 11 & 67 \\
        LARP~\cite{wang2024larp}& 173M & 632M & 8192 & 1024 & 20 & 5.1  & \underline{57} \\
        DeRA (Ours) & 174M & 632M & 8192 & 1024 & \underline{15} & \textbf{4.1}  & \textbf{50} \\
        \bottomrule
        \end{tabular}
    }
    \vspace{-0.1in}
    \caption{Comparison of video reconstruction and generation results. Results are grouped by the type of generative models. ``$\downarrow$'' indicates lower is better. \textbf{Bold} and \underline{underlining} denote the best and the second best performance, respectively.}
    \label{tab:comp}
\end{table*}

\subsection{Symmetric Alignment-Conflict Projection}
\label{sec:sacp}
To balance the gradient conflicts between $\mathcal{L}_{\text{align}}^{a}$ and $\mathcal{L}_{\text{align}}^{m}$ during joint optimization, we propose Symmetric Alignment-Conflict Projection (SACP), which symmetrically projects the gradient from one stream onto the other during training to achieve a balanced update direction.

As outlined in Alg.~\ref{alg:SACP}, at each optimization step, we first compute the gradients $\mathbf{g}_a$ and $\mathbf{g}_m$ with respect to the encoder parameters $\theta_\mathcal{E}$. When a gradient conflict is detected, we compute the projection coefficients $c_a$ and $c_m$ by normalizing their dot product $\mathbf{s}$ with the $\ell_2$ norm of the opposing alignment gradient. To maintain numerical stability and computational efficiency, these coefficients are detached from the computation graph by $\text{stopgrad}(\cdot)$ and treated as first-order constants rather than variables~\cite{zhang2022does}. Finally, these coefficients are used to re-weight the alignment losses, ensuring that the final optimization step follows a consensus direction beneficial to both alignment objectives:
\begin{equation}
\label{eq:sacp-proj}
\mathcal{L}_{\text{re}}^{a}, \mathcal{L}_{\text{re}}^{m} =
\text{SACP}(\mathcal{L}_{\text{align}}^{a}, \mathcal{L}_{\text{align}}^{m}, \theta_\mathcal{E}, \varepsilon_\text{stab}),
\end{equation}
where the constant $\varepsilon_\text{stab}$ is empirically set to $1\mathrm{e}{-8}$ to ensure numerical stability and prevent division-by-zero errors~\cite{kingma2014adam}.

\begin{algorithm}[!tb]
\caption{Symmetric Alignment-Conflict Projection}
\label{alg:SACP}
\begin{algorithmic}
\REQUIRE Encoder parameters $\theta_\mathcal{E}$, alignment losses $\mathcal{L}_{\text{align}}^{a}$, $\mathcal{L}_{\text{align}}^{m}$, stability constant $\varepsilon_\text{stab}$.
\ENSURE Reformulated loss $\mathcal{L}_{\text{re}}^{a}, \mathcal{L}_{\text{re}}^{m}$.
\STATE $\mathbf{g}_a \leftarrow \nabla_{\theta_E}\mathcal{L}_{\text{align}}^{a}$ \quad $\mathbf{g}_m \leftarrow \nabla_{\theta_E}\mathcal{L}_{\text{align}}^{m}$
\STATE $\mathbf{s} \leftarrow \langle \mathbf{g}_a, \mathbf{g}_m \rangle$
\IF{$s < 0$}
  \STATE $c_a \leftarrow \mathrm{stopgrad}\!\left(\dfrac{\mathbf{s}}{\|\mathbf{g}_m\|_2+\varepsilon_\text{stab}}\right)$
  \STATE $c_m \leftarrow \mathrm{stopgrad}\!\left(\dfrac{\mathbf{s}}{\|\mathbf{g}_a\|_2+\varepsilon_\text{stab}}\right)$
  \STATE $\mathcal{L}_{\text{re}}^{a} \leftarrow \mathcal{L}_{\text{align}}^{a} - c_a\,\mathcal{L}_{\text{align}}^{m}$
  \STATE $\mathcal{L}_{\text{re}}^{m} \leftarrow \mathcal{L}_{\text{align}}^{m} - c_m\,\mathcal{L}_{\text{align}}^{a}$
\ELSE
  \STATE $\mathcal{L}_{\text{re}}^{a} \leftarrow \mathcal{L}_{\text{align}}^{a}$; \quad $\mathcal{L}_{\text{re}}^{m} \leftarrow \mathcal{L}_{\text{align}}^{m}$
\ENDIF
\end{algorithmic}
\end{algorithm}

\subsection{Autoregressive Visual Generation}
As mentioned in Sec.~\ref{sec:archi}, DeRA tokenizes the video into a sequence of discrete tokens $\mathbf{y}=(\mathbf{y}_1, ...,\mathbf{y}_L)$. This representation enables us to model video generation as a next-token prediction task, akin to autoregressive (AR) language modeling. A decoder-only transformer~\cite{touvron2023llama,sun2024autoregressive} is trained to maximize the conditional log-likelihood of the target token $\mathbf{y}_{i}$ with cross-entropy loss:
\begin{equation}
\label{eq:argen}
\max _{\theta} \sum_{i=1}^{L} \log p({\mathbf{y}}_{i} \mid \mathbf{c}, \mathbf{y}_{1: i-1};{\theta}),
\end{equation}
where $\mathbf{c}$ denotes the condition for class-conditional video generation, $p$ and $L$ represent the probability distribution and the length of $\mathbf{y}$, and $\theta$ is the learnable parameter of the AR model. During inference, we autoregressively sample a sequence of tokens according to the model likelihood, then the predicted tokens are passed to decoder $\mathcal{D}$ for decoding.

\section{Experiments}
\label{sec:experiments}

\subsection{Experimental Settings}
\textbf{Datasets.} We follow previous work~\cite{wang2024larp,li2025learning} to train DeRA on UCF-101~\cite{soomro2012dataset} and Kinetics-600 (K600)~\cite{carreira2018short}. We evaluate both reconstruction fidelity and generation quality on these datasets. In all experiments, we use 16-frame clips with a resolution of $128\times128$ for training and evaluation.

\vspace{0.02in}
\noindent\textbf{Implementation details.}
In all experiments, the patch sizes are set to $t=4$ and $p=8$, following~\cite{wang2024omnitokenizer,wang2024larp}. We set the number of appearance and motion tokens to 256 and 768, respectively. The hidden dimension is set to 768, the latent dimension is 16, and the codebook size is 8192. Unless otherwise specified, we use the DINOv3 ViT-B/16~\cite{simeoni2025dinov3} as the image foundation model following~\cite{yu2024representation,wu2025representation}, and InternVideo2-B/14~\cite{wang2024internvideo2} as the video foundation model. The alignment loss weights for the appearance $\mathcal{L}_{\text{align}}^{a}$ and motion $\mathcal{L}_{\text{align}}^{m}$ are set to $\lambda_a = 1.0$ and $\lambda_m = 0.5$, respectively.

\vspace{0.02in}
\noindent\textbf{Video generation.}
We employ a LLaMA-style~\cite{touvron2023llama,sun2024autoregressive} transformer, following~\cite{wang2024larp}. For class-conditional generation on the UCF-101 dataset, a [CLS] token is prepended to the sequence as a class-conditional prompt, while for frame prediction on the K600 dataset, a [SEP] token is used to separate the context and target sequences. We adopt the Fréchet Video Distance (FVD)~\cite{unterthiner2018towards} as the main metric for all evaluations. During sampling, we set the CFG scale to 1.2. For ablation studies, we train the tokenizer for 75 epochs, and use the 343M-parameter generator following LARP~\cite{wang2024larp}.

\begin{figure}[!tb]
\centering
\includegraphics[width=0.48\textwidth]{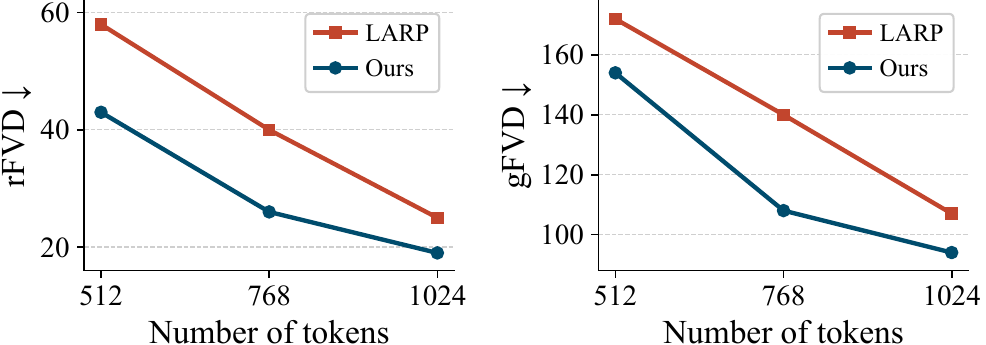}
\vspace{-0.2in}
\caption{The reconstruction and generation comparisons using different token budgets.}
\label{fig:token_budget}
\end{figure}

\begin{figure*}[!tb]
\centering
\includegraphics[width=\textwidth]{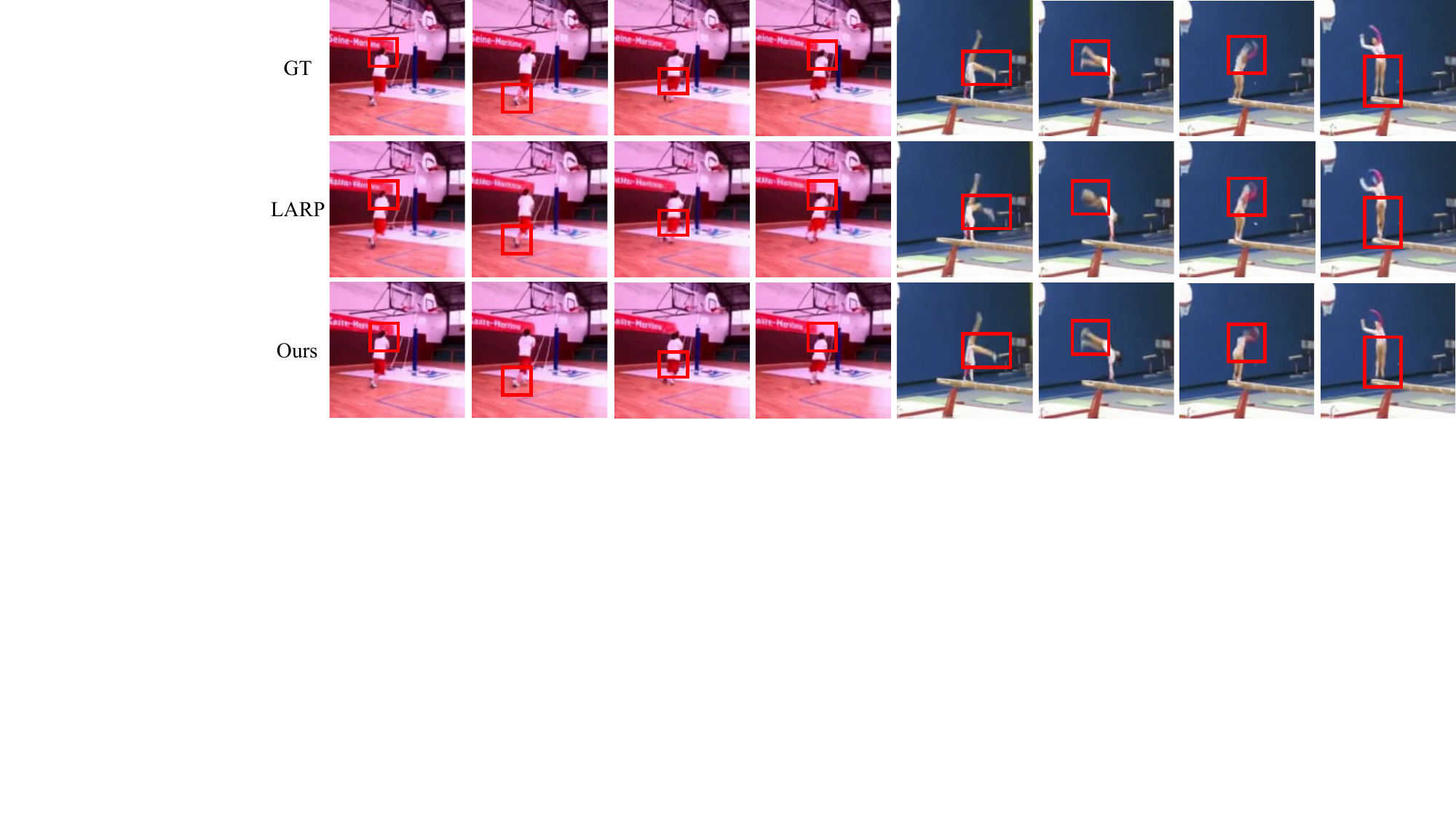}
\vspace{-0.2in}
\caption{Comparison of video reconstruction results between LARP~\cite{wang2024larp} and our method on UCF-101~\cite{soomro2012dataset} dataset.}
\label{fig:rec_results}
\end{figure*}

\subsection{Video Tokenization and Generation}
We first evaluate the video tokenization ability of DeRA on the UCF-101 dataset. As shown in Tab.~\ref{tab:comp}, DeRA achieves superior reconstruction performance with the same compression rate and parameter. Additionally, as illustrated in Fig.~\ref{fig:token_budget}, DeRA requires fewer tokens to achieve comparable performance to LARP~\cite{wang2024larp}, which is crucial for improving the efficiency of practical AR generation.

To evaluate the video generation performance, we compare DeRA with other state-of-the-art video generative models, including diffusion-based methods, masked language modeling (MLM) methods, and autoregressive (AR) methods. Following the settings in~\cite{wang2024omnitokenizer,wang2024larp,li2025learning}, we use the UCF-101 class-conditional generation and the K600 frame prediction benchmark, where the first 5 frames of a 16-frame video clip are provided to predict the next 11 frames. As shown in Tab.~\ref{tab:comp}, DeRA establishes a new state-of-the-art performance, outperforming all other methods on the UCF-101 and K600 datasets, including the closed-source MAGVIT-v2~\cite{yu2023language}, demonstrating clear gains in temporal coherence and perceptual fidelity. Moreover, DeRA delivers consistent performance gains under various token budgets in Fig.~\ref{fig:token_budget}. This highlights the robustness and compression efficiency of DeRA, validating that our decoupled representations are both more expressive and more compact.

\vspace{-0.05in}
\subsection{Ablation Study}
\label{sec:abl}

\noindent\textbf{Effects of decoupled alignment.} To investigate the impact of decoupled alignment, we conduct experiments that align either the appearance or motion stream independently. As shown in Tab.~\ref{tab:abl_decouple_align}, aligning a single stream improves both rFVD and gFVD over the non-aligned baseline, indicating that representation alignment provides effective supervision even in isolation~\cite{leng2025repa}. Furthermore, aligning both streams simultaneously achieves the best performance across all metrics, confirming their complementarity and emphasizing the importance of leveraging distinct foundation models for decoupled alignment. Interestingly, aligning the appearance stream with DINOv3~\cite{simeoni2025dinov3} yields larger performance gains than aligning the motion stream with InternVideo2~\cite{wang2024internvideo2}, suggesting that image foundation models currently offer stronger and more stable supervisory signals than their video counterparts.

\begin{table}[!tb]
\centering
\begin{tabularx}{\linewidth}{l Y Y Y Y}
\toprule
\textbf{Method} & \textbf{PSNR}$\uparrow$ & \textbf{LPIPS}$\downarrow$ & \textbf{rFVD}$\downarrow$ & \textbf{gFVD}$\downarrow$ \\
\midrule
Base & 28.06 & 0.0813 & 24.47 & 112\\
\quad + DINOv3 & 28.31 & 0.0791 & 20.14 & 98\\
\quad + IV2 & 28.16 & 0.0804 & 22.64 & 107\\
\quad + DINOv3 + IV2 & \textbf{28.49} & \textbf{0.0777} & \textbf{18.83} & \textbf{94}\\
\bottomrule
\end{tabularx}
\vspace{-0.1in}
\caption{Ablation study on the decoupled alignment effect. IV2 denotes InternVideo2~\cite{wang2024internvideo2}.}
\label{tab:abl_decouple_align}
\end{table}

\begin{figure*}[!tb]
\centering
\includegraphics[width=\textwidth]{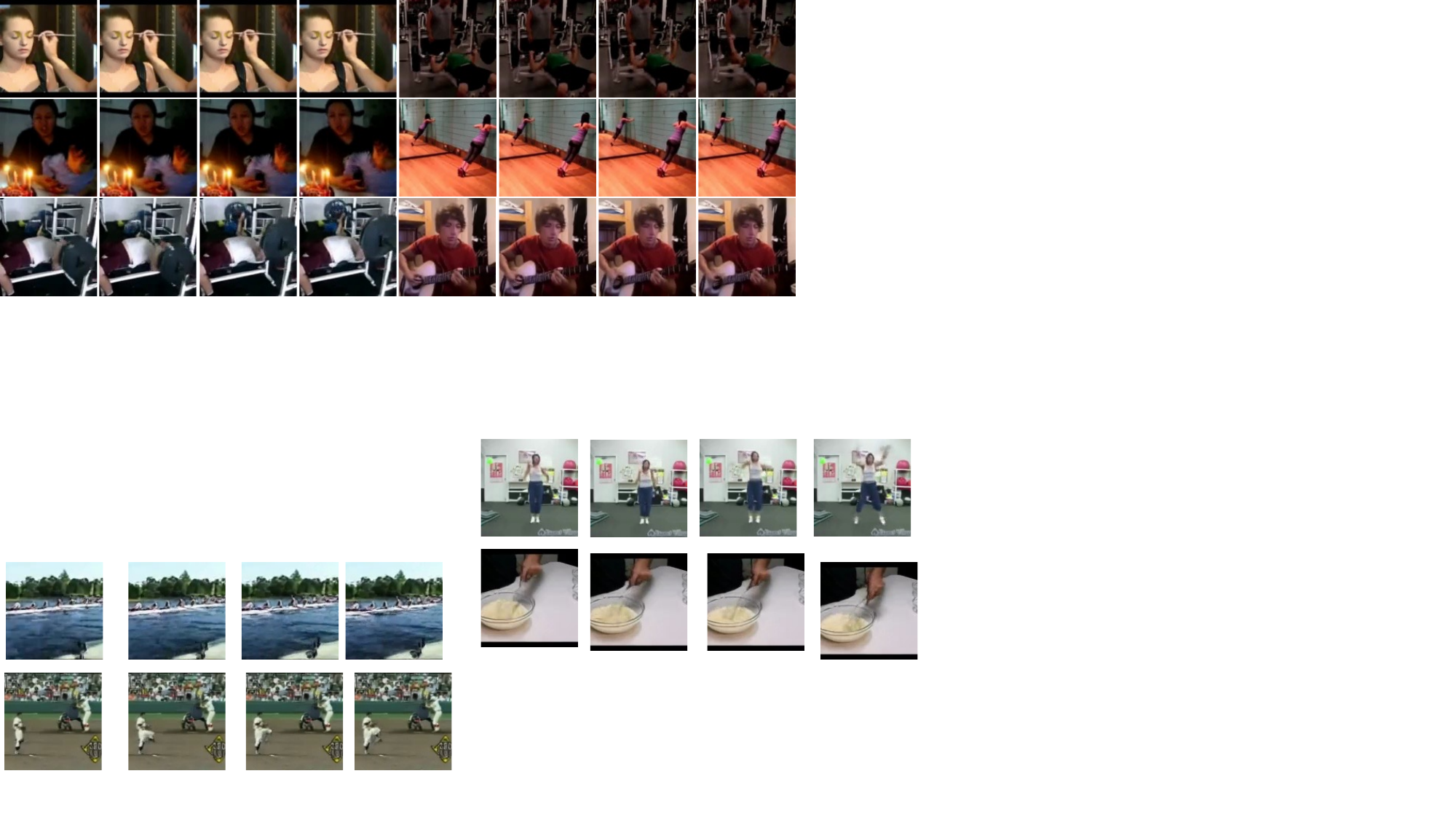}
\vspace{-0.2in}
\caption{Class-conditional video generation comparison on UCF-101~\cite{soomro2012dataset} dataset. Left: LARP~\cite{wang2024larp}; right: Ours.}
\label{fig:class_results}
\end{figure*}

\begin{table}[!tb]
\centering
\renewcommand{\arraystretch}{1.0}
\begin{tabularx}{\linewidth}{l Y Y}
\toprule
\textbf{Method} & \textbf{rFVD}$\downarrow$ & \textbf{gFVD}$\downarrow$ \\
\midrule
w/o alignment &24.48 & 112\\
\midrule
DINOv2-B/14 & 23.98 & 109\\
DINOv3 ViT-B/16 & \textbf{20.14} & \textbf{98}\\ 
\midrule
VideoMAEv2 & 24.14 & 110\\
InternVideo2-B/14 & \textbf{22.64} & \textbf{107}\\
\bottomrule
\end{tabularx}
\vspace{-0.1in}
\caption{Ablation study on the aligned latent space.}
\label{tab:target_model}
\end{table}
\vspace{-0.1in}

\begin{table}[!tb]
\centering
\begin{tabularx}{\linewidth}{l c c c c}
\toprule
\textbf{Method} & \textbf{$\lambda_a$} & \textbf{$\lambda_m$} & \textbf{rFVD}$\downarrow$ & \textbf{gFVD}$\downarrow$ \\
\midrule
Base & - & - & 24.48 &112\\
\midrule
DINOv3 + InternVideo2     & 0.5 & 0.5 & 20.01 & 101\\
DINOv3 + InternVideo2     & 1 & 0.5 & \textbf{18.83} & \textbf{94}\\
DINOv3 + InternVideo2     & 0.5 & 1 & 19.65 & 99\\
DINOv3 + InternVideo2     & 1 & 1 & 18.94 & 95\\
\midrule
Two-stage training & 1 & 0.5 & 19.52 & 99\\
\bottomrule
\end{tabularx}
\vspace{-0.1in}
\caption{Ablation study on alignment settings. Two-stage training denotes only aligning the image patches at the first stage, then aligning both image and video patches at the second stage.}
\label{tab:aliset}
\end{table}

\vspace{0.15in}
\noindent\textbf{Effects of aligned latent space.} We investigate the impact of using different vision foundation models as aligned targets in Tab.~\ref{tab:target_model}. The results show that 1) representation alignment consistently yields substantial improvements across all aligned models, enhancing both reconstruction and generation performance; and 2) more advanced models outperform their earlier counterparts by providing stronger supervision, \eg, DINOv3~\cite{simeoni2025dinov3} brings 3.84 rFVD and 11 gFVD gains compared to DINOv2~\cite{oquab2023dinov2}. 

\begin{table}[!tb]
\centering
\begin{tabularx}{\linewidth}{l Y Y Y}
\toprule
\textbf{Method} & \textbf{Weight} & \textbf{rFVD}$\downarrow$ & \textbf{gFVD}$\downarrow$ \\
\midrule
Base & - & 19.81 & 97\\
\quad + Soft Loss & 0.5 & 19.42 & 96\\
\quad + Soft Loss & 1 & 20.12 & 103\\
\quad + SACP & - & \textbf{18.83} & \textbf{94}\\
\bottomrule
\end{tabularx}
\vspace{-0.1in}
\caption{Comparison on resolving multi-alignment gradient conflict. ``Base'' denotes our DeRA without conflict regularization.}
\label{tab:conflict}
\end{table}

\begin{table}[!tb]
\centering
\renewcommand{\arraystretch}{1.0}
\begin{tabularx}{\linewidth}{l Y Y Y}
\toprule
\textbf{Method} & \textbf{Depth} & \textbf{rFVD}$\downarrow$ & \textbf{gFVD}$\downarrow$ \\
\midrule
w/o alignment & - & 24.48 & 112\\
\midrule
DINOv3 & 6 & 21.34 & 103\\
DINOv3 & 8 & 21.68 & 103\\
DINOv3 & 10 & 20.79 & 101\\
DINOv3 & 12 & \textbf{20.14} & \textbf{98}\\ 
\midrule
InternVideo2 & 6 & 23.45 & 110\\
InternVideo2 & 8 & 24.17 & 110\\
InternVideo2 & 10 & 22.96 & 108\\
InternVideo2 & 12 & \textbf{22.64} & \textbf{107}\\
\midrule
DINOv3 + InternVideo2 & 8 & 20.12 & 99\\
DINOv3 + InternVideo2 & 10 & 19.14 & 97\\
DINOv3 + InternVideo2 & 12 & \textbf{18.83} & \textbf{94}\\
\bottomrule
\end{tabularx}
\vspace{-0.1in}
\caption{Ablation study on the alignment depth. ``Depth’’ denotes the encoder layer index whose features will be aligned.}
\label{tab:abl_model}
\vspace{-0.1in}
\end{table}

\noindent\textbf{Effects of alignment strategy.} Tab.~\ref{tab:aliset} shows the ablation study on the weights of alignment loss $\mathcal{L}_{\text{align}}^{a}$ and $\mathcal{L}_{\text{align}}^{m}$. Among the evaluated values, $\lambda_a=1$ and $\lambda_m=0.5$ yields the best performance on both rFVD and gFVD. We also implement a two-stage training strategy, which first aligns the appearance stream for 20 epochs and then both streams for 55 epochs. As shown in the final row, it offers no significant gains over our standard one-stage joint training.

\begin{figure*}[!tb]
\centering
\includegraphics[width=\textwidth]{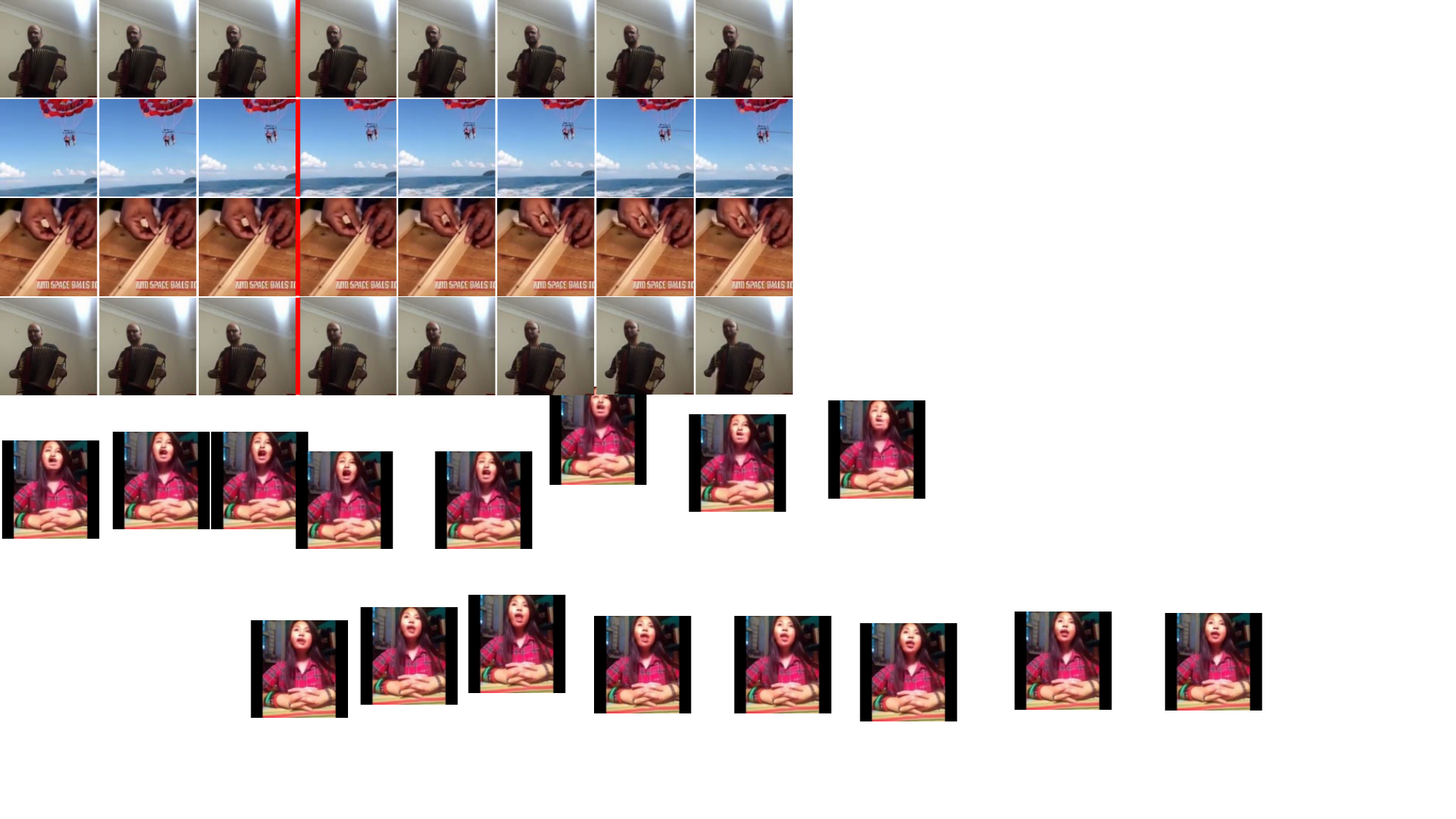}
\vspace{-0.25in}
\caption{Comparison of frame prediction results between LARP~\cite{wang2024larp} (Rows 1, 3) and ours (Rows 2, 4) on Kinetics-600~\cite{carreira2018short} dataset. The left 3 frames are conditioned frames, while the right 5 frames are generated.}
\label{fig:frame_results}
\end{figure*}

\begin{figure*}[!tb]
\centering
\includegraphics[width=\textwidth]{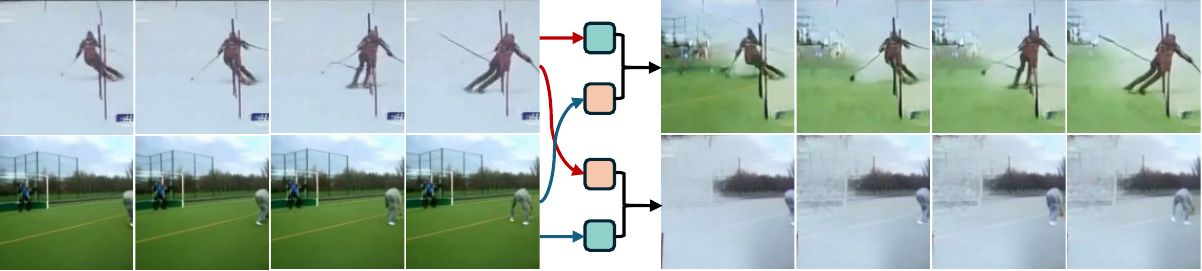}
\vspace{-0.23in}
\caption{Visualization of the token swapping results by DeRA: benefited from the decoupled latent space, DeRA can do zero-shot video editing by swapping the appearance and motion tokens of two distinct videos. }
\label{fig:token_swapping}
\end{figure*}

\vspace{0.02in}
\noindent\textbf{Effects of SACP.} We evaluate the effectiveness of SACP in resolving conflicts between the gradients of two alignment objectives in Tab.~\ref{tab:conflict}. Soft Loss, which leverages an optimization loss to penalize the negative dot product of the two conflicting gradients, is chosen as a baseline. We find that Soft Loss is sensitive to penalty weight, \ie, setting the loss weight to 0.5 brings improvements while using 1.0 degrades performance. Comparatively, SACP consistently improves both reconstruction and generation performance by enforcing conflict-free updates without introducing new tunable hyperparameters and with negligible overhead.

\vspace{0.02in}
\noindent\textbf{Effects of alignment depth.} We further study the effect of alignment layer selection, as shown in Tab.~\ref{tab:abl_model}. For both DINOv3 and InternVideo2, performance consistently improves when aligning with features from deeper layers, with the final encoder layer (Depth = 12) yielding the best results.

\subsection{Visualizations}
\noindent\textbf{Video reconstruction.} We visualize the reconstruction results by LARP~\cite{wang2024larp} and our DeRA in Fig.~\ref{fig:rec_results}. Our method performs significantly better than the baseline, capturing both high-frequency spatial textures and maintaining temporal stability even during rapid motion like 
gymnastics.

\vspace{0.02in}
\noindent\textbf{Class-conditional video generation.} We present class-conditional video generation results in Fig.~\ref{fig:class_results}. DeRA synthesizes class-faithful videos with fine textures, stable structures, and coherent long-range motion. The decoupled alignment provides the AR generator with a well-structured and more semantic latent space, reducing identity drift and preserving spatial detail across frames.

\vspace{0.02in}
\noindent\textbf{Video frame prediction.} We show the frame prediction results in Fig.~\ref{fig:frame_results}. The vertical red line marks the boundary between the conditioned context and the predicted frames. We use 5 frames as input to predict the following 11 frames, forming a 16-frame video clip. The results demonstrate that DeRA generates continuations that are not only plausible but also consistent and temporally stable, faithfully extrapolating the motion from the context.

\vspace{0.02in}
\noindent\textbf{Token swapping.} To further explore the decoupling effect, we conduct the token swapping experiment to visualize the semantic independence of appearance and motion tokens. Surprisingly, the synthesized results in Fig.~\ref{fig:token_swapping} inherit the texture, identity, and background from the appearance source while following the dynamics of the motion source. This outcome demonstrates that the appearance and motion tokens of DeRA are independently controllable, offering direct qualitative evidence of semantic factorization and validating the effectiveness of our decoupled alignment design.

\section{Conclusion}
\label{sec:conclusion}
This work introduces DeRA, a novel 1D video tokenizer that explicitly decouples spatial and temporal representation learning in video tokenization. DeRA adopts a dual-stream encoder architecture, where the appearance stream captures spatial details under the guidance of an image foundation model, while the motion stream learns temporal dynamics with the supervision of a video foundation model. To stabilize the multi-objective optimization, we further propose a Symmetric Alignment-Conflict Projection (SACP) module, which mitigates gradient conflicts through adaptive gradient reformulation. Extensive experiments demonstrate that DeRA achieves competitive results on both reconstruction and autoregressive generation benchmarks. 

In the future, we will explore the pathway to scale DeRA for better performance and compression efficiency, as well as extending DeRA toward semantic video autoencoders~\cite{zheng2025diffusion} for more expressive tokenization.

{
    \small
    \bibliographystyle{ieeenat_fullname}
    \bibliography{main}
}

% WARNING: do not forget to delete the supplementary pages from your submission..
% \input{X_suppl}

\end{document}